\documentclass[a4paper,fleqn]{cas-dc}

\usepackage[authoryear]{natbib}

\def\tsc#1{\csdef{#1}{\textsc{\lowercase{#1}}\xspace}}
\tsc{WGM}
\tsc{QE}
\tsc{EP}
\tsc{PMS}
\tsc{BEC}
\tsc{DE}

\usepackage{cuted}
\usepackage{longtable}
\usepackage{parskip}

\usepackage{xurl}

\begin{document}
\let\WriteBookmarks\relax
\def\floatpagepagefraction{1}
\def\textpagefraction{.001}
\shorttitle{Lung CT Benchmark}
\shortauthors{Nils Neukirch et~al.}

\title [mode = title]{Foundation Models vs. Radiomics for Lung Computed Tomography: A Benchmark of Feature Extractors, Classification Heads, and Segmentation Choices}

\author[1]{Nils Neukirch}[prefix=,
                          orcid=0009-0000-4036-6144]

\ead{nils.neukirch@uol.de}
                        
\author[2]{Martin Maurer}[prefix=Prof. Dr. Dr.]
\ead{martin.maurer@uol.de}

\author[1]{Nils Strodthoff}[prefix=Prof. Dr.,orcid=0000-0003-4447-0162]
\ead{nils.strodthoff@uol.de}
\cormark[1]

\affiliation[1]{organization={Division AI4Health, Carl von Ossietzky Universität Oldenburg},
                addressline={Ammerländer Heerstraße 114-118}, 
                city={Oldenburg},
                citysep={},
                postcode={26129}, 
                state={Niedersachsen},
                country={Germany}}

\affiliation[2]{organization={University Institute for Diagnostic and Interventional Radiology, Klinikum Oldenburg AöR},
                addressline={Rahel-Straus-Straße 10}, 
                city={Oldenburg},
                citysep={},
                postcode={26133}, 
                state={Niedersachsen},
                country={Germany}}

\cortext[1]{Corresponding author}

\begin{abstract}
Radiomics is the established approach for CT-based lung cancer phenotyping, yet comparisons with foundation models rarely isolate contributions of feature extractor, classification head, and segmentation choice, or test cross-cohort robustness. We benchmark five feature extractors (Curia, Curia-2, DINOv3, Radiomics~2D, Radiomics~3D), seven classification heads (TabPFN, TabICL, XGBoost, CatBoost, Random Forest, logistic regression, Ridge), and three segmentation regimes on five tasks: tumor volume and stage classification, 2-year survival prediction, histology classification, and age prediction. Models are trained on LUNG1 ($n=338$) and evaluated on an internal test set ($n=84$) and the external LUNG2 cohort ($n=211$), with worst-case cross-cohort performance as the primary metric. The dominant design factor is task-dependent: segmentation drives volume and stage classification, while classifier choice drives survival, histology, and age prediction. Radiomics is competitive for tumor volume, tumor stage and survival (partly due to label-derivation effects for the former); Curia variants reach comparable peak scores for survival; DINOv3 falls slightly short across tasks. Patch and slice aggregation have negligible impact. We recommend Curia with tumor segmentation and a CatBoost head as a safe default, achieving the best mean rank across the three primary clinical tasks, though task-specific selection consistently outperforms any cross-task default. When tumor delineations are unavailable, Curia-2 with lung segmentation and logistic regression offers a competitive alternative. All pipelines use a two-stage design suited to small cohort sizes where end-to-end fine-tuning would risk overfitting.
\end{abstract}


\maketitle

\section{Introduction}
\label{sec:intro}
Lung cancer is the leading cause of cancer-related mortality worldwide, and CT-based phenotyping (from tumor staging to outcome prediction) plays a central role in clinical decision-making \citep{Bray2024-lg}. For more than a decade, \emph{radiomics} has been the dominant paradigm for building such predictors: hand-engineered features such as first-order statistics, shape descriptors, and texture matrices are extracted from a delineated tumor region and fed to a tabular classifier \citep{Parmar2015-lr}. Despite a large body of published radiomic signatures, the field is confronted with well-documented reproducibility concerns and limited cross-cohort generalizability \citep{Welch2019-ep,Raptis2024-dt}.

Recently, imaging foundation models trained on large medical image datasets via self-supervised or contrastive objectives have emerged as an alternative representation source \citep{Pai2024-ko,Dancette2025Curia}. By encoding generic visual structure rather than task-specific features, these models yield embeddings that can be reused across tasks with a simple downstream head.

Yet, most published comparisons between radiomics and foundation-model embeddings are narrow in scope: they fix the segmentation (usually the tumor mask), fix the downstream classifier, and report a single cohort pair. It therefore remains unclear (a) whether foundation-model features require a tumor mask at all, (b) which downstream head best exploits high-dimensional embeddings, and (c) how robust either approach is when moving to an unseen external cohort. In this work, we address these questions through a systematic factorial benchmark.

Our benchmark specifically targets the small-sample regime with $\mathcal{O}(100)$ samples with global labels, which is challenging to address within approaches trained in an end-to-end manner. Both typical radiomics solutions and the foundation-model-based approaches explored here circumvent this challenge through a two-stage process: in a first step (predefined/pretrained) features are extracted, which are then processed by a tabular classifier in a second step.

More specifically, our contributions are:
\begin{enumerate}
    \item A factorial benchmark over five feature extractors, seven classification heads, three segmentation choices, and (for 2D features) patch- and slice-aggregation variants, evaluated on five clinically relevant tasks.
    \item An evaluation protocol that exposes cross-cohort robustness rather than internal performance alone.
    \item Concrete practical recommendations covering both task-specific configurations and a \emph{safe-default} pipeline achieving the best mean rank across the three primary clinical tasks without task-specific tuning, including a \emph{no-tumor-mask} alternative for settings where expert delineations are unavailable.
\end{enumerate}

\section{Related Work}
\label{sec:related}

\textbf{Radiomics on LUNG1/LUNG2.}
The NSCLC-Radiomics (LUNG1) \citep{Aerts2019-aq} and NSCLC-Radiogenomics (LUNG2) \citep{Bakr2018} cohorts have been widely used as a train/test pair for radiomic outcome prediction. Reported AUCs for staging tasks range from 0.61 to 0.76, and C-indices for overall survival range from 0.60 to 0.73, with variability depending on feature selection and validation strategy \citep{Parmar2015-lr,Chaddad2017-rr,Shi2019-ty,Welch2019-ep,Yang2019-jc,Scalco2024-sf}.

\textbf{Deep learning on lung CT.}
Several works have combined or compared radiomics and CNN-based features on the same cohorts. \citet{Haarburger2019-dg} report a C-index of 0.623 for a radiomics+DL fusion versus 0.585 for DL alone; \citet{Braghetto2022-qk} find that radiomics and radiomics+DL both reach an AUC of 0.67 for survival, with DL alone at 0.63.

\textbf{Imaging foundation models.}
Self-supervised pretraining on large radiology datasets yields general-purpose embeddings that transfer across tasks. Curia \citep{Dancette2025Curia} is a CT-specific model pretrained contrastively on large-scale radiology data; the Harvard cancer-imaging foundation model \citep{Pai2024-ko} similarly demonstrates that such representations can encode clinically relevant cancer biomarkers. Curia-2 \citep{saporta2026curia2scalingselfsupervisedlearning} extends this approach with an improved self-supervised pre-training strategy. General purpose vision foundation models such as DINOv3 \citep{siméoni2025dinov3} can serve as non-medical baseline.

\textbf{Scalable 3D classification with adapted foundation models.}
\citet{liu2026revisiting2dfoundationmodels} propose AnyMC3D, a 2D-to-3D framework that fine-tunes a frozen 2D backbone with LoRA adapters and aggregates slice embeddings via learned attention pooling, demonstrating strong gains over naive frozen-feature baselines across 12 diverse tasks. Our work shares the slice-wise 2D encoder design but differs in three key aspects. First, AnyMC3D operates with up to 50\,000 training volumes per task and more than half of its benchmark tasks rely on private datasets, placing it in a fundamentally different data regime from our small-sample setting ($n=338$ training patients on fully public cohorts). Second, its evaluations are conducted within a single held-out split per task, without cross-cohort external validation. Third, no comparison to radiomics baselines is provided, leaving open whether the gains over frozen-feature pooling extend to classical hand-engineered features.

\textbf{Position of this work.}
Across the studies above, two dimensions are consistently underexplored: cross-cohort robustness and systematic ablation of the full pipeline. We address both by holding the data fixed (LUNG1/LUNG2) and varying five extractors, seven classification heads, three segmentation choices, and two aggregation stages, enabling direct attribution of performance differences to individual design decisions. Crucially, every configuration is scored on both an internal and an independent external cohort, with worst-case performance across the two as our primary metric, a robustness criterion absent from most prior work. This design also allows us to directly compare foundation-model features against radiomics baselines under identical conditions.

\section{Methods}
\label{sec:methods}

\begin{figure*}[ht]
    \centering
    \includegraphics[width=0.7\textwidth]{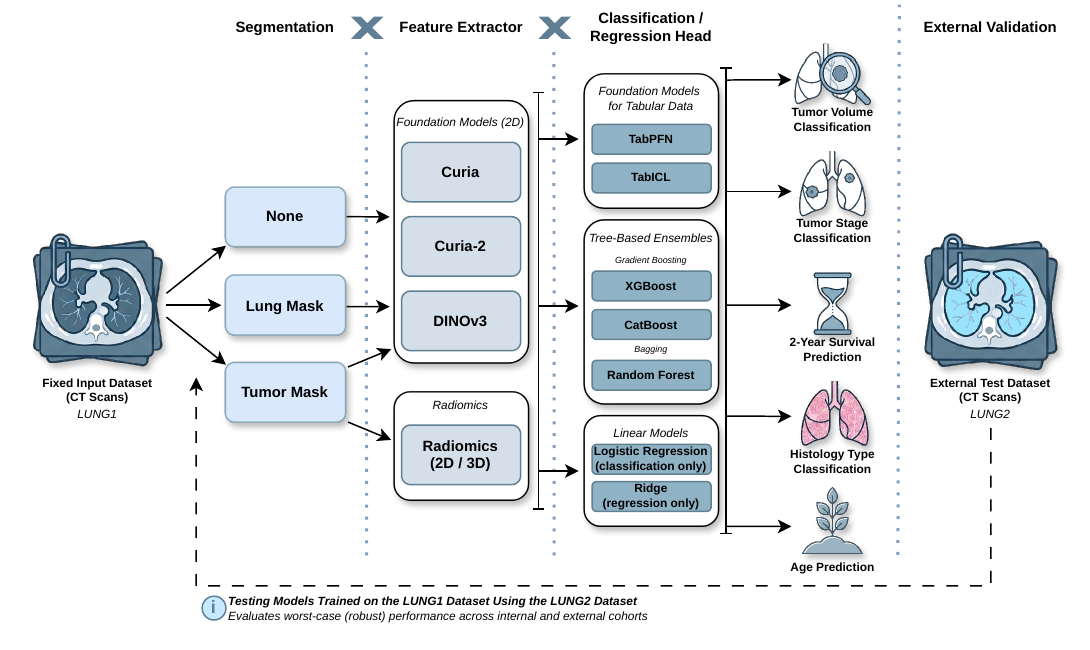}
    \caption{\textbf{Benchmark overview.} For each CT volume we vary three axes:
    (i)~the \emph{segmentation} provided to the feature extractor (no mask, lung mask or tumor mask);
    (ii)~the \emph{feature extractor}: a foundation model (Curia, Curia-2, DINOv3) or hand-engineered radiomics (2D, 3D);
    (iii)~the \emph{classification head}: a tabular foundation model (TabPFN, TabICL), a linear model (logistic regression / Ridge), a bagged tree (Random Forest), or a gradient-boosted tree (XGBoost / CatBoost).
    Training is performed on the LUNG1 training split; evaluation is performed on the LUNG1 internal test split and on the full external LUNG2 cohort. Slice and patch aggregation axes are not shown as they show negligible impact on performance. Google Nano Banana 2 was used to generate the icons for this figure.}
    \label{fig:schematic}
\end{figure*}

Figure~\ref{fig:schematic} gives an overview of the benchmark. For each CT volume, we vary five design axes: the segmentation provided to the feature extractor, the feature extractor itself, and the downstream classification or regression head. We evaluate every resulting pipeline configuration on an internal (LUNG1) and an external (LUNG2) cohort. For clarity, slice and patch aggregation are omitted from the figure as they show negligible impact on performance.

\subsection{Datasets and tasks}
\label{sec:methods:data}
We use two public lung-CT cohorts from The Cancer Imaging Archive:

\textbf{LUNG1 / NSCLC-Radiomics}: 422 NSCLC patients. Split into a training set (338 patients, 80\%) and an internal test set (84 patients, 20\%), generated by stratified 5-fold cross-validation with random seed 42; the same folds are shared across all feature extractors and tasks, ensuring direct comparability.

\textbf{LUNG2 / NSCLC-Radiogenomics}: 211 NSCLC patients. Used \emph{only} as an external test cohort, never seen during training.

We evaluate five prediction tasks:

\textbf{Tumor volume classification} (binary classification, AUC): binarised at the within-cohort median volume of 39{,}039\,mm$^3$, yielding a balanced split of 210 patients per class. Tumor volume is a primary component of TNM staging and a strong independent predictor of prognosis; automated volume stratification can streamline staging workflows and support eligibility screening for clinical trials.

\textbf{Tumor stage classification} (binary classification, AUC): T1/T2 (early stage, $n=249$) vs.\ T3/T4 (advanced stage, $n=172$); 1 patient with missing stage excluded. Tumor stage is the principal determinant of treatment strategy in NSCLC, guiding decisions between surgical resection, radiotherapy, and systemic therapy; non-invasive stage prediction from pre-treatment CT is especially relevant in centres where biopsy access is limited.

\textbf{2-year survival prediction} (binary classification, AUC): patients who died within 2 years form the negative class ($n=251$, 59.8\%); 2 patients with missing survival data excluded. Two-year survival is a standard clinical endpoint in NSCLC outcome research; imaging-based survival prediction could help identify high-risk patients who may benefit from intensified surveillance or adjuvant treatment.

\textbf{Histology type classification} (binary classification, AUC): squamous cell carcinoma ($n=152$, 40.0\%) vs.\ all other histologies ($n=228$, 60.0\%); 42 patients with missing annotations excluded. Distinguishing squamous from non-squamous histology is directly treatment-relevant, as certain first-line regimens (e.g.\ bevacizumab-based chemotherapy) are contraindicated in squamous NSCLC; imaging-based histology prediction offers a non-invasive complement to tissue biopsy when re-biopsy carries procedural risk.

\textbf{Age prediction} (regression, MAE): continuous age prediction; $n=400$ patients with valid labels (mean $68.0\pm10.1$ years, range 33.7--91.7\,years); 22 patients excluded. Imaging-derived age estimates may serve as a surrogate for biological or functional age, complementing calendar age in assessing treatment tolerance and fitness for aggressive therapy protocols.

Applying the LUNG1-derived thresholds and label definitions to LUNG2 induces several covariate and label-distribution shifts: a subset of tasks has substantially fewer valid labels on LUNG2 (notably tumor volume and 2-year survival), and the LUNG2 cohort is more strongly imbalanced for tumor volume, tumor stage, and histology type; age labels are available for all patients and match the LUNG1 distribution closely. These shifts are quantified in Appendix~\ref{app:methods:dist_shift} and should be kept in mind when interpreting cross-cohort results.

CT volumes are processed in their native acquisition geometry (no resampling to a common voxel spacing prior to foundation-model extraction), with intensity handling and mask resampling differing by extractor; radiomics uses a separate IBSI-compliant pipeline. The full per-extractor preprocessing details are given in Appendix~\ref{app:methods:preprocessing}.

\subsection{Feature extractors}
\label{sec:methods:features}
We compare two families of feature extractors. As \emph{foundation models} we use \textbf{Curia} \citep{Dancette2025Curia} and its successor \textbf{Curia-2} \citep{saporta2026curia2scalingselfsupervisedlearning}, both CT-specific models pretrained with self-supervised learning on large-scale radiology data, together with \textbf{DINOv3} \citep{siméoni2025dinov3}, a general-purpose (non-medical) vision model included as a baseline for the value of domain-specific pretraining. All three operate slice-wise on 2D axial slices and produce patch-level embeddings that are combined into a patient-level representation by the two-stage aggregation of Section~\ref{sec:methods:agg}. As hand-engineered baselines we use \emph{radiomics} features computed with PyRadiomics (IBSI-compliant) in two variants: \textbf{Radiomics 2D} (features computed per slice and aggregated across slices) and \textbf{Radiomics 3D} (features computed once on the whole tumor volume). Architectural specifications, pretraining corpora, input resolutions and patch grids, and the full radiomics extraction settings (feature classes, resampling, discretization) are given in Appendix~\ref{app:methods:features}.

\subsection{Segmentation regimes}
\label{sec:methods:seg}
For the foundation-model extractors we vary which mask (if any) is applied to the CT before feature extraction:

\textbf{Tumor mask}: the ground-truth tumor delineation provided with the dataset.

\textbf{Lung mask}: a whole-lung mask obtained from the ground-truth delineations provided with LUNG1; for LUNG2, lung masks were computed automatically using the CLIP-Driven Universal Model \citep{liu2023clip}.

\textbf{No segmentation}: the full CT is passed to the extractor.

Radiomics always operates on the tumor mask by definition.

\subsection{Aggregation for 2D features}
\label{sec:methods:agg}
Foundation models that operate slice-wise produce a feature vector per patch and per slice. To obtain a single vector per patient we aggregate in two stages: \textbf{patch aggregation} (\texttt{mean} or \texttt{cov}, i.e.\ covariance pooling~\citep{dooms2026covariance}) across patches within a slice, followed by \textbf{slice aggregation} (\texttt{max}, \texttt{mean}, or \texttt{median}) across slices. 3D extractors and radiomics output a single vector per patient, so these axes are not applicable (\texttt{n/a}). Because of TabPFN's constraints, we do not use covariance pooling for TabPFN (Appendix~\ref{app:methods:agg}).

\subsection{Prediction heads}
\label{sec:methods:heads}
We train seven tabular models on the resulting feature vectors: TabPFN \citep{hollmann2022tabpfn}, TabICL \citep{qu2025tabicltabularfoundationmodel}, XGBoost \citep{chen2016xgboost}, CatBoost \citep{prokhorenkova2018catboost}, Random Forest, logistic regression (classification tasks), and Ridge regression (age prediction). All heads use TabArena / AutoGluon default hyperparameters; the exact settings and the early-stopping protocol used for the boosting methods are listed in Appendix~\ref{app:methods:heads}.

\subsection{Evaluation protocol}
\label{sec:methods:eval}
Each configuration $c$ = (\text{features}, \text{segmentation}, \text{patch\_agg}, \text{slice\_agg}, \text{head}) is trained on the LUNG1 training split and scored on (i) the LUNG1 internal test split and (ii) the full LUNG2 cohort. The primary metric is AUC for classification and mean absolute error (MAE) for age prediction; 95\% confidence intervals are obtained by bootstrap resampling of the test-set predictions (Appendix~\ref{app:methods:bootstrap}).
 
We define the robustness score for a configuration $c$ as the
conservative estimate across cohorts:
\[
    \mathrm{robust\_score}(c) = \min\bigl(\mathrm{score}_{\mathrm{LUNG1}}(c),\;
    \mathrm{score}_{\mathrm{LUNG2}}(c)\bigr)
\]
for AUC tasks (higher is better), and
\[
    \mathrm{robust\_score}(c) = \max\bigl(\mathrm{score}_{\mathrm{LUNG1}}(c),\;
    \mathrm{score}_{\mathrm{LUNG2}}(c)\bigr)
\]
for the MAE task (lower is better). Both cases thus return the \emph{worst-case} performance across cohorts and thereby penalizes configurations that win on one cohort by overfitting to it.
For brevity we continue to use the notation $\mathrm{min\_score}(c)$ in tables and figures, with the understanding that for age prediction the relevant quantity is the maximum MAE across cohorts.

\section{Results}
\label{sec:results} 
\subsection{Cross-cohort generalization: LUNG1 vs.\ LUNG2}
\label{sec:results:scatter}
\begin{figure*}[!ht]
    \centering
    \includegraphics[width=\textwidth]{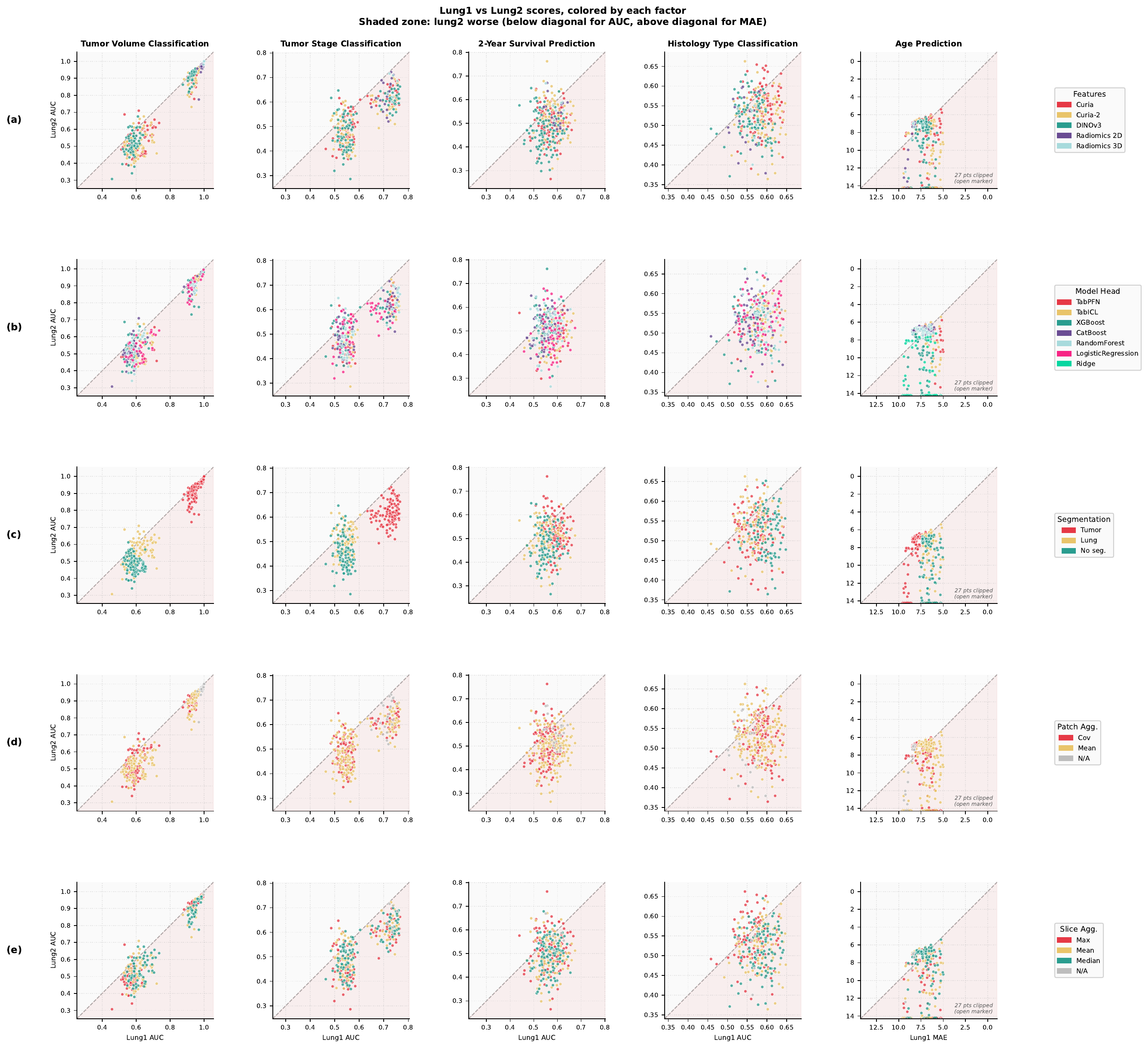}
    \caption{\textbf{LUNG1 vs.\ LUNG2 performance for all configurations.}
    Each point is one (features, segmentation, patch agg, slice agg, head) configuration; columns are the five tasks; points below the dashed diagonal have a performance drop on the external cohort. The five panels color the same data by different pipeline factors: (a)~feature extractor, (b)~classification head, (c)~segmentation choice, (d)~patch aggregation and (e)~slice aggregation. Panel~(a):  Radiomics~2D/3D configurations cluster at high LUNG1 and LUNG2 AUC for tumor volume classification, reflecting the label-derivation artifact; for survival prediction and histology type classification, foundation-model configurations span a wider range and reach higher peak scores than radiomics. Panel~(b): Ridge regression produces a distinct cluster of high MAE outliers for age (visible as points far above the diagonal), confirming it as the dominant source of variance for that task; for classification tasks no single head separates cleanly. Panel~(c): Tumor-mask configurations concentrate in the upper-right of the first three classification panels,  indicating both higher absolute performance and better cross-cohort consistency; no-segmentation points cluster in the lower-left, with many falling below the diagonal. Panels~(d) and~(e): Patch and slice aggregation points overlap almost completely across all tasks, confirming negligible influence on both LUNG1 and LUNG2 performance.}
    \label{fig:scatter_combined}
\end{figure*}

Figure~\ref{fig:scatter_combined} shows, for each task, every configuration's LUNG1 score against its LUNG2 score, with panels~(a) to~(e) coloring the same set of points by different pipeline factors. Configurations on the diagonal generalize perfectly; configurations below the diagonal lose performance on the external cohort.

Excluding the most extreme $5\%$ of configurations on each side, the trimmed mean LUNG1\,$\to$\,LUNG2 AUC drop is $0.06$ for tumor volume classification, $0.08$ for tumor stage, $0.07$ for 2-year survival, and $0.05$ for histology type classification. For age prediction, the trimmed mean gap of $1.2$~years is modest, suggesting that age-related parenchymal patterns transfer well across cohorts once extreme outliers are excluded. 

\subsection{What matters most for robust performance?}
\label{sec:results:strips}
To quantify which pipeline factor most influences robust performance, we compute for each factor the mean $\Delta$ (best minus worst level) across all configurations that vary only by that factor. The complete tables showing each configuration are provided in the supplementary material. Table~\ref{tab:fi_summary} reports these values per task; larger values indicate a greater impact on performance. Strip plots for all five tasks, showing the full distribution of cross-cohort scores by factor level, are provided in Appendix~\ref{app:strips_remaining}.

\begin{table*}[htbp]
\centering
\caption{Mean $\Delta$ (best\,--\,worst) per factor and task. Larger values indicate greater influence on the outcome.}
\label{tab:fi_summary}
\begin{tabular}{l r r r r r}
\toprule
Factor & Tumor Vol. & T-Stage & Surv. 2yr & Histology & Age (MAE) \\
\midrule
Features & 0.096 & 0.093 & 0.110 & 0.090 & 3.754 \\
Classifier & 0.106 & 0.113 & 0.129 & 0.107 & 9.451 \\
Segmentation & 0.418 & 0.176 & 0.089 & 0.072 & 4.654 \\
Patch Agg. & 0.046 & 0.051 & 0.060 & 0.045 & 2.345 \\
Slice Agg. & 0.071 & 0.072 & 0.085 & 0.069 & 2.736 \\
\bottomrule
\end{tabular}
\end{table*}

The dominant factor is strongly task-dependent. Segmentation is the most influential factor for tumor volume and tumor stage classification ($\Delta = 0.42$ and $0.18$, respectively), while classifier choice dominates for 2-year survival, histology type classification, and age prediction ($\Delta = 0.13$, $0.11$, and $9.45$, respectively). Feature extractor has moderate influence across all tasks ($\Delta = 0.09$--$0.11$ for classification tasks). Patch and slice aggregation have negligible impact across all tasks ($\Delta < 0.09$ for all classification tasks), indicating these design decisions can be made freely without meaningful performance loss.

\subsection{Best configuration per factor level}
\label{sec:results:top_per_factor}
\begin{figure*}[!ht]
    \centering
    \includegraphics[width=\textwidth]{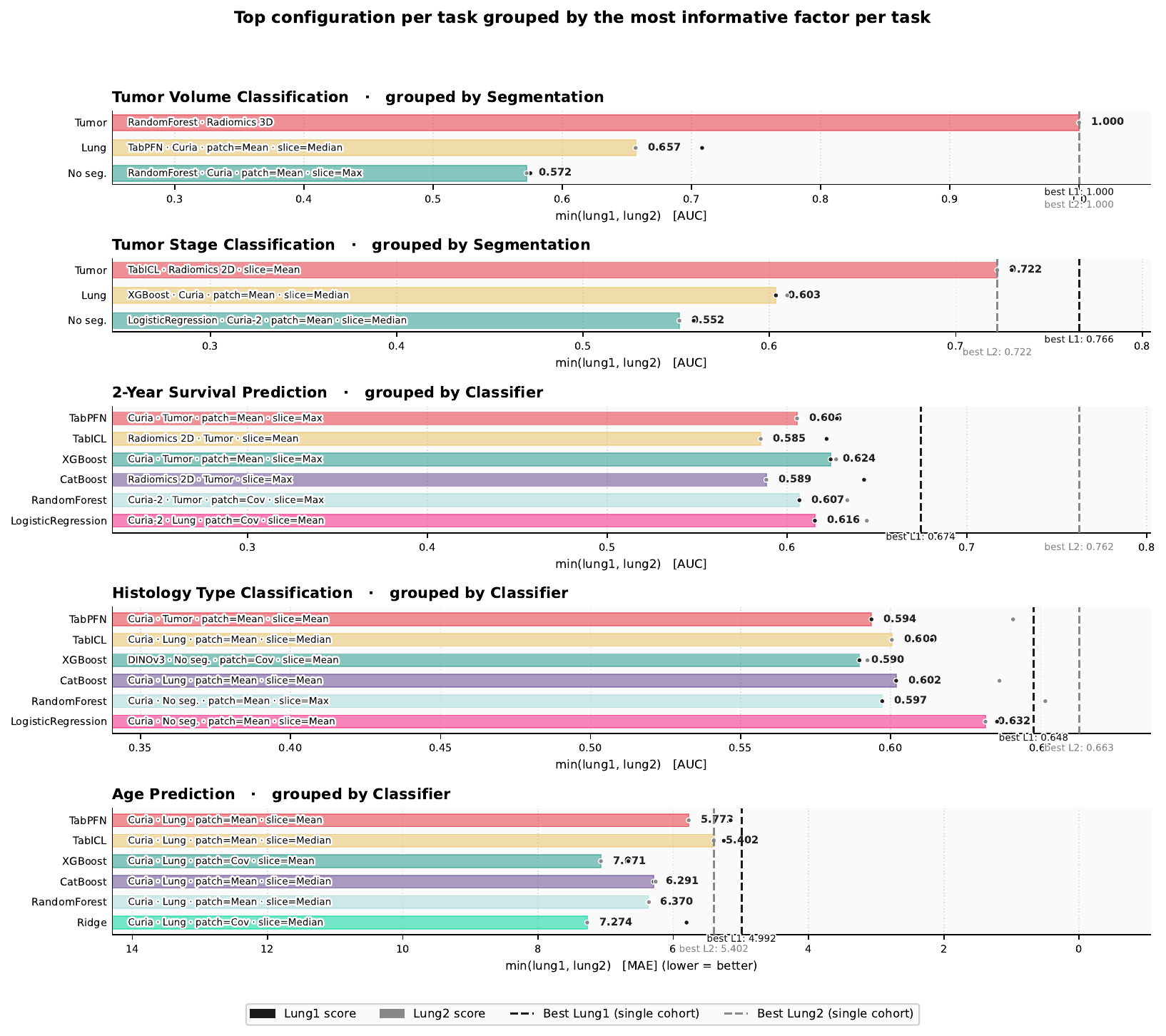}
    \caption{\textbf{Best configuration per factor level.} Bar length = robust score: $\mathrm{min}(\mathrm{LUNG1}, \mathrm{LUNG2})$ AUC for classification tasks; $\mathrm{max}(\mathrm{LUNG1}, \mathrm{LUNG2})$ MAE for age prediction (i.e.\ worst-case error across cohorts; note the reversed x-axis). Filled circles show the LUNG1 (black) and LUNG2 (gray) scores of that same best-robust configuration; dotted lines indicate the single-cohort best scores (the highest LUNG1 score and highest LUNG2 score achievable with that factor level, potentially from different configurations). For each task the factor with the largest mean $\Delta$ is shown (cf.\ Table~\ref{tab:fi_summary}); bars correspond to that factor's levels. Segmentation choice drives the greatest performance spread for tumor volume and tumor stage classification ($\Delta = 0.42$ and $0.18$, respectively) while classifier choice is most influential for 2-year survival ($\Delta = 0.13$), histology type classification ($\Delta = 0.11$) and age prediction ($\Delta = 9.45$). Comparing the dotted lines to the dots reveals how much performance is sacrificed by requiring cross-cohort robustness. For some histology configurations, the LUNG2 score noticeably exceeds the LUNG1 score, suggesting those embeddings transfer particularly well to the external cohort.}
    \label{fig:top_per_factor_all}
\end{figure*}

Figure~\ref{fig:top_per_factor_all} visualises, the most influential pipeline factor for each task as measured by mean $\Delta$ (cf.\ Table~\ref{tab:fi_summary}). For that factor, each bar represents one factor level and shows the best-configuration robustness score (the robust, both-cohort metric) achieved by any pipeline using that level; the black and gray dots show the LUNG1 and LUNG2 scores of that same configuration. In addition, dotted lines indicate the single-cohort best scores: the highest LUNG1 score and the highest LUNG2 score achievable with that factor level, potentially from different configurations. The gap between a dotted line and the corresponding dot quantifies how much performance is sacrificed by requiring good generalization across both cohorts simultaneously.

\subsection{Best overall configurations across clinical tasks}
\label{sec:results:overall}

Table~\ref{tab:best_configs_all_t_stage_binary_survived_2yr_histology} lists the top~15 pipeline configurations ranked by mean $\mathrm{min\_score}$ AUC across the three primary clinical tasks: tumor stage classification, 2-year survival prediction, and histology type classification. Tumor volume classification is excluded because its binary label is derived from the same tumor segmentation used by the radiomics extractor, conferring an inherent advantage to tumor-mask configurations (see Section~\ref{sec:discussion:limitations}); age prediction is excluded as it is not a standard clinical endpoint in lung cancer phenotyping.

The overall best configuration is Curia with tumor segmentation, mean patch aggregation, mean slice aggregation, and a CatBoost head, achieving AUC~0.668 for tumor stage (rank~21), AUC~0.543 for 2-year survival prediction (rank~61), and AUC~0.582 for histology type classification (rank~20). While it is possible to use this single configuration across all tasks, it does not reach the peak performance achievable by task-specific configurations (e.g.\ rank~21 on tumor stage vs.\ a peak of rank~1 for that task), underscoring how different the underlying prediction problems are. The top~15 is dominated by tumor-mask configurations; notably, overall rank~13 (Curia-2, lung segmentation, cov patch aggregation, mean slice aggregation, logistic regression) demonstrates that competitive performance is achievable without any tumor delineation, which is practically relevant when expert contours are unavailable.

\begin{table*}[t]
\centering
\small
\setlength{\tabcolsep}{4pt}
\caption{
  Top 15 configurations (all variants), sorted by mean rank across tasks: tumor stage classification, 2yr. survival prediction, histology type classification. Each cell shows rank\,(AUC). Lower rank is better (1\,=\,best within task).
}
\label{tab:best_configs_all_t_stage_binary_survived_2yr_histology}
\begin{tabular}{r l c r r r}
\toprule
\# & Configuration & \textbf{Rank}$\downarrow$ & \textbf{T-Stage} & \textbf{Surv. 2yr} & \textbf{Histology} \\
\midrule
    1 & Curia · Tumor · p=Mean · s=Mean · CatBoost & 34.0 & 21 (0.668) & 61 (0.543) & 20 (0.582) \\
    2 & Curia-2 · Tumor · p=Cov · s=Max · RandomForest & 39.0 & 84 (0.599) & 3 (0.607) & 30 (0.576) \\
    3 & Curia · Tumor · p=Mean · s=Mean · TabPFN & 41.7 & 29 (0.651) & 86 (0.532) & 10 (0.594) \\
    4 & Radiomics 2D · Tumor · s=Max · RandomForest & 45.0 & 54 (0.622) & 57 (0.546) & 24 (0.580) \\
    5 & DINOv3 · Tumor · p=Cov · s=Mean · RandomForest & 47.0 & 58 (0.619) & 27 (0.568) & 56 (0.565) \\
    6 & Curia-2 · Tumor · p=Mean · s=Median · XGBoost & 47.3 & 39 (0.635) & 29 (0.568) & 74 (0.555) \\
    7 & Curia · Tumor · p=Mean · s=Mean · XGBoost & 47.7 & 37 (0.636) & 37 (0.561) & 69 (0.559) \\
    8 & Radiomics 3D · Tumor · TabPFN & 49.7 & 10 (0.686) & 11 (0.587) & 128 (0.538) \\
    9 & Curia-2 · Tumor · p=Cov · s=Median · RandomForest & 50.3 & 5 (0.699) & 41 (0.557) & 105 (0.545) \\
    10 & Curia · Tumor · p=Cov · s=Mean · RandomForest & 51.7 & 22 (0.667) & 66 (0.542) & 67 (0.560) \\
    11 & Radiomics 2D · Tumor · s=Max · CatBoost & 53.7 & 41 (0.632) & 8 (0.589) & 112 (0.542) \\
    12 & Curia · Tumor · p=Cov · s=Max · RandomForest & 55.3 & 96 (0.587) & 24 (0.572) & 46 (0.569) \\
    13 & Curia-2 · Lung · p=Cov · s=Mean · LogisticRegression & 58.3 & 145 (0.535) & 2 (0.616) & 28 (0.576) \\
    14 & DINOv3 · Tumor · p=Mean · s=Max · RandomForest & 58.7 & 23 (0.663) & 103 (0.521) & 50 (0.568) \\
    15 & Curia · Tumor · p=Cov · s=Median · CatBoost & 60.3 & 35 (0.639) & 85 (0.532) & 61 (0.562) \\
\bottomrule
\end{tabular}
\end{table*}

\section{Discussion}
\label{sec:discussion}
 
\subsection{Which configurations work where}
\label{sec:discussion:configs}

The factor-influence analysis (Table~\ref{tab:fi_summary} and Figure~\ref{fig:top_per_factor_all}) shows that the dominant design choice is strongly task-dependent. For tumor volume classification, segmentation is by far the most influential factor ($\Delta = 0.42$): providing the ground-truth tumor mask yields substantially higher robust AUC than a lung mask or no mask. This result must be interpreted carefully, however: the volume label was itself derived from the same tumor segmentation used by radiomics, giving tumor-mask configurations a near-circular advantage. For tumor stage classification, segmentation also dominates ($\Delta = 0.18$), consistent with tumor size being a defining staging criterion. For 2-year survival, histology type classification and age prediction, classifier choice has the greatest impact ($\Delta = 0.13$, $\Delta = 0.11$ and $9.45$). Across all tasks, patch and slice aggregation have negligible influence ($\Delta < 0.10$ in all cases for classification tasks), indicating that these design decisions can be made freely without meaningful performance loss.

Based on these findings, we recommend \textbf{Curia with tumor mask, mean patch aggregation, mean slice aggregation, and a CatBoost head} as a safe default: it achieves the best mean rank across tumor stage, 2-year survival prediction, and histology type classification (Table~\ref{tab:best_configs_all_t_stage_binary_survived_2yr_histology}). When tumor delineations are unavailable, \textbf{Curia-2 with lung segmentation and a logistic regression head} (rank~13 overall) offers a competitive mask-free alternative, demonstrating that foundation-model embeddings can transfer across cohorts even without explicit spatial masking. Notably, this safe default does not match the peak performance achievable when configurations are selected per task, highlighting that the prediction tasks are sufficiently heterogeneous to warrant task-specific pipeline choices when performance is critical.

\subsection{Foundation models vs.\ radiomics}
\label{sec:discussion:fm_vs_rad}

Radiomics features dominate the top configurations for tumor volume classification, a finding partly explained by the label-derivation artifact noted above. For 2-year survival, radiomics achieves the highest median robust AUC, but Curia and Curia-2 reach equally competitive peak scores, suggesting that the two approaches are broadly comparable on this task. DINOv3 configurations generally fall slightly short of the Curia variants, possibly reflecting the absence of CT-specific pretraining and/or its lower input resolution (Section~\ref{sec:discussion:limitations}). For tumor stage, radiomics 2D and 3D achieve the highest median robust AUC among all extractors (cf.\ Figure~\ref{fig:strip_tstage}), consistent with tumor size being the primary staging criterion and radiomics shape features directly encoding size information. The peak robust score (0.722) is reached by a TabICL configuration using Radiomics~2D with the tumor mask, while Curia variants approach this ceiling when the tumor mask is provided. For histology type classification, the advantage shifts to foundation models: Curia achieves the highest median robust AUC (best: 0.632), followed closely by Curia-2 and DINOv3, while radiomics shows lower medians for this task (cf.\ Figure~\ref{fig:strip_histology}).
Notably, for several Curia-2 histology configurations the LUNG2 score exceeds the LUNG1 score; while this is consistent with the embeddings transferring well to the external cohort, it may also partly reflect the strong class-imbalance shift in LUNG2, which is skewed towards non-squamous tumors (83.4\%, vs.\ 60.0\% in LUNG1, cf.\ Appendix~\ref{app:methods:dist_shift}), making the majority class easier to predict and inflating LUNG2 AUC independently of representation quality.

\subsection{Role of the prediction head}
\label{sec:discussion:head}

The influence of the classification head is strongly task-dependent. For age prediction, head choice is the dominant factor ($\Delta = 9.45$\,years), driven by Ridge regression performing poorly relative to all other heads: its median robust MAE substantially exceeds that of TabPFN, TabICL, tree-based, and boosting methods. TabICL achieves the best robust MAE (5.4\,years).

For classification tasks, no single head dominates consistently across all tasks. Notably, linear heads (logistic regression) are competitive with non-linear alternatives when used with foundation-model embeddings (see Appendix~\ref{app:strips_remaining}), suggesting that the embeddings themselves carry sufficient structure for linear separability, an observation consistent with the representation-learning literature.

\subsection{Role of segmentation}
\label{sec:discussion:seg}

Providing spatial context (tumor or lung mask) to the feature extractor generally outperforms using the full unmasked CT, as shown in the strip plots (Appendix~\ref{app:strips_remaining}). The tumor mask achieves the highest median performance on most tasks, with two exceptions: histology type classification and age prediction, where the lung mask performs better. When tumor delineations are unavailable, the lung mask offers a practical fallback that recovers a meaningful share of the performance gain. 

Overall, the findings align with radiological expert knowledge. For tumor stage classification, stage is defined by tumor size and local extent, so directing the extractor to the tumor mask focuses features on the morphological properties most directly encoding the staging criterion. For 2-year survival prediction, intra-tumoral heterogeneity (reflecting necrosis, central low density, and irregular boundaries) is considered a prognostic imaging biomarker, and tumor-mask features capture these properties more cleanly than full-image features. For histology type classification and age prediction, however, the lung context appears informative beyond the tumor itself: squamous cell carcinomas tend to arise centrally with associated atelectasis or post-obstructive consolidation, while adenocarcinomas present peripherally with ground-glass components, and the lung mask preserves these larger anatomical relationships. Age-related parenchymal changes (including emphysema, interstitial patterns, and vascular calcifications) are distributed throughout the lung rather than confined to the tumor, which may explain why the lung mask outperforms the tumor mask for age prediction.

These findings stress the necessity for robust zero-shot or few-shot segmentation pipelines that can produce reliable segmentations even for small-scale datasets. Similarly, it highlights the possibility to build lightweight aggregation modules that implement data-driven pooling mechanisms acting on top of unpooled foundation model representation to reduce the dependence on segmentation masks.
\subsection{Limitations}
\label{sec:discussion:limitations}

This benchmark covers two cohorts only; the conclusions may not generalize to other scanners, institutions, or NSCLC patient populations. LUNG2 is substantially smaller than LUNG1 (211 vs.\ 422 patients) and has only 63 patients with valid 2-year survival labels, limiting statistical power for that task. All benchmarks are conducted on a single organ site (lung); whether the findings generalize to other anatomical regions or cancer types remains an open question. The three foundation-model extractors share a patch size of 16 and an embedding dimension of 768, but differ in effective input resolution: Curia and Curia-2 operate at $512{\times}512$ (1{,}024 patches per slice), whereas DINOv3 operates at its native $256{\times}256$ (256 patches per slice). DINOv3's lower performance may therefore reflect either the absence of CT-specific pretraining, the reduced spatial resolution, or both; our design does not disentangle these factors. Ground-truth tumor segmentations in both cohorts originate from a single annotator; inter-annotator variability is not modelled. No domain adaptation is applied between cohorts. Finally, the tumor-volume task is partly circular: the binary label is derived from the same tumor mask that radiomics uses as input, giving radiomics an inherent advantage that would be absent in any real-world deployment setting where the tumor volume is not known in advance.

\section{Conclusion}
In this work, we presented a first comprehensive, comparative study on radiomics vs.\ foundation model features in the small-sample regime, using lung cancer as a case study. Our study provides explicit guidance for the practitioner on design choices, ranging from feature extractors over segmentation masks and aggregation methods to prediction heads, highlighting setups that promise high performance and good generalization.

\clearpage
\section*{Data availability}
The LUNG1 (NSCLC-Radiomics) dataset \citep{Aerts2019-aq} and LUNG2 (NSCLC-Radiogenomics) dataset \citep{Bakr2018} are publicly available through The Cancer Imaging Archive at \url{https://www.cancerimagingarchive.net/collection/nsclc-radiomics} and \url{https://www.cancerimagingarchive.net/collection/nsclc-radiogenomics}, respectively.

\section*{Code availability}
The code developed in this study is publicly available to ensure reproducibility and to facilitate further research. It is accessible at: \url{https://github.com/AI4HealthUOL/lung-ct-benchmarking}

\section*{Declaration of generative AI and AI-assisted technologies in the manuscript preparation process}
During the preparation of this work the authors used Anthropic Claude (claude-sonnet-4-6) as a writing assistant to help structure the manuscript and draft selected sections. Google Nano Banana 2 was used to generate the icons for the benchmark overview (Fig.~\ref{fig:schematic}); the rest of the diagram was created by hand.

After using these tools, the authors reviewed and edited the content as needed and take full responsibility for the content of the published article.

\section*{Declaration of competing interests}
The authors declare that they have no known competing financial interests or personal relationships that could have appeared to influence the work reported in this paper.

\section*{Acknowledgements}
This work was supported by the Bundesministerium für Forschung, Technologie und Raumfahrt (BMFTR) under Project INGVER, grant number 01KX2419.

\appendix
\onecolumn

\section{Methodological and Implementation Details}
\label{app:methods_details}
This appendix collects the technical descriptions referenced from the Methods section (Section~\ref{sec:methods}): image preprocessing, feature-extractor specifications, aggregation and classifier implementation details, cross-cohort label-distribution shifts, and the bootstrap confidence-interval procedure.

\subsection{Image preprocessing}
\label{app:methods:preprocessing}
CT volumes were processed in their native acquisition geometry: no spatial resampling to a common voxel spacing was applied prior to foundation-model feature extraction. The foundation-model extractors operate slice-wise on every axial slice of the volume; spatial restriction to a tumor or lung region (where applicable) is imposed downstream at the patch-aggregation stage (Section~\ref{sec:methods:agg}) rather than by cropping or resampling the input. Intensity handling differs by extractor. For DINOv3, each axial slice is windowed to the fixed Hounsfield Unit (HU) range $[-1000, 400]$\,HU, linearly rescaled to 8-bit $[0, 255]$, replicated across three channels, and then resized and normalized using the model's native ImageNet preprocessing transform. For Curia and Curia-2, raw HU values are passed directly to the model's accompanying image processor, which applies its own per-slice $z$-score normalization. Segmentation masks, when used, are resampled with nearest-neighbor interpolation to the CT grid whenever their geometry differs from the corresponding CT volume. Radiomics features use a separate preprocessing pipeline (isotropic resampling to $1{\times}1{\times}1$\,mm and fixed-bin-width discretisation), described in Appendix~\ref{app:methods:features}.

\subsection{Feature extractor details}
\label{app:methods:features}

\subsubsection*{Foundation models}
\textbf{Curia} \citep{Dancette2025Curia} is a CT/MRI vision foundation model built on a DINOv2 ViT-B backbone (86M parameters), pretrained with self-supervised learning on a large-scale real-world radiology corpus of 150{,}000 cross-sectional imaging exams (130\,TB). It operates slice-wise (2D): each axial slice is resized to $512{\times}512$ and passed as raw Hounsfield-Unit values (no windowing) to the model's accompanying image processor, which performs per-slice intensity normalization internally. With a patch size of 16 this yields a $32{\times}32$ grid of 1{,}024 patch-level embeddings of dimension 768, plus a class token. Patient-level representations are obtained via the two-stage aggregation described in Section~\ref{sec:methods:agg}.

\textbf{Curia-2} \citep{saporta2026curia2scalingselfsupervisedlearning} is a follow-up to Curia that improves the self-supervised pre-training strategy and representation quality for radiological data, while leaving the underlying architecture, embedding dimension, patch grid, and inference procedure identical to Curia (Section~\ref{sec:methods:agg}).

\textbf{DINOv3} \citep{siméoni2025dinov3} is a general-purpose vision foundation model pretrained via self-distillation on large-scale natural image datasets (ViT-B/16 backbone). It is applied slice-wise to CT volumes: each axial slice is windowed to $[-1000, 400]$\,HU, rescaled to 8-bit, replicated across three channels, and resized to $256{\times}256$ using the model's native ImageNet preprocessing. With a patch size of 16 this produces a $16{\times}16$ grid of 256 patch-level embeddings of dimension 768; the 4 register tokens are discarded so the patch layout matches Curia, and patient-level representations are obtained via the same two-stage aggregation (Section~\ref{sec:methods:agg}). No CT-specific fine-tuning was applied, making it a natural baseline for assessing the value of domain-specific pretraining.

\subsubsection*{Radiomics}

\textbf{Radiomics 2D}: features computed per slice (force-2D extraction on slices containing at least one masked voxel) and aggregated across slices.

\textbf{Radiomics 3D}: features computed once on the whole tumor volume. 

Features from all seven PyRadiomics feature classes were extracted on the original (unfiltered) image: 14 shape, 18 first-order, 22 GLCM, 16 GLRLM, 16 GLSZM, 14 GLDM, and 5 NGTDM, yielding 105 features per extraction in both the 2D (per-slice, then aggregated to a single patient vector via median across slices) and 3D regimes. Prior to extraction, images were resampled to an isotropic voxel spacing of $1{\times}1{\times}1$\,mm using B-spline interpolation (nearest-neighbor for masks), and intensities were discretized with a fixed bin width of 25\,HU (IBSI recommendation for CT). No additional filter banks (e.g.\ wavelet or LoG) and no per-image intensity normalization were applied.

\subsection{Two-stage aggregation: implementation notes}
\label{app:methods:agg}
For patch aggregation with TabPFN, only \texttt{mean} pooling is used, as covariance pooling yields 4\,096-dimensional vectors that exceed TabPFN's internal feature limit of 2\,000.

\subsection{Classifier hyperparameters}
\label{app:methods:heads}
All heads use TabArena/AutoGluon default hyperparameters: Random Forest with 300 trees and a maximum of 15\,000 leaf nodes; XGBoost with learning rate 0.1 and \texttt{gbtree} booster; CatBoost with learning rate 0.05 and balanced class weights; logistic regression with $\ell_2$ penalty ($C=1.0$, \texttt{lbfgs} solver, balanced class weights); Ridge with $\alpha=1.0$; and TabPFN~v2.6 with 32 estimators.
The sole exception is the number of boosting rounds for XGBoost and CatBoost, which is determined via early stopping (patience: 50 rounds, maximum: 10\,000) on a held-out inner validation set (20\,\% of the training fold for radiomics; first split of a 5-fold inner cross-validation for foundation-model features).

\subsection{Cross-cohort label distribution shifts}
\label{app:methods:dist_shift}
When the LUNG1-derived thresholds and label definitions are applied to LUNG2, notable distributional differences arise that affect external evaluation. For tumor volume classification, only 143 of 211 LUNG2 patients have a valid segmented volume; furthermore, the LUNG2 median volume (8{,}249\,mm$^3$) is substantially lower than the LUNG1 threshold (39{,}039\,mm$^3$), producing a strongly imbalanced split (117 low / 26 high, 81.8\% / 18.2\%). For 2-year survival prediction, only 63 of 211 LUNG2 patients have non-missing labels ($n=36$ deceased, $n=27$ survived), limiting the statistical power of the LUNG2 evaluation for this task. For tumor stage and histology type classification, LUNG2 is skewed towards early-stage (82.1\% T1/T2) and non-squamous tumors (83.4\%), which differs markedly from the LUNG1 distribution; these covariate shifts should be kept in mind when interpreting cross-cohort results. Age labels are available for all 211 LUNG2 patients (mean $68.0\pm10.0$ years), matching the LUNG1 distribution closely.

\subsection{Bootstrap confidence intervals}
\label{app:methods:bootstrap}
We report 95\% confidence intervals computed via the empirical (pivotal) bootstrap with 1{,}000 resamples drawn with replacement from the test-set predictions; resampling was not stratified by class label.

\clearpage
\section{Strip Plots: All Tasks}
\label{app:strips_remaining}

Strip plots for all five prediction tasks. Each dot is one (features, segmentation, patch agg, slice agg, head) configuration, positioned by $\mathrm{min\_score}$ on the x-axis and grouped on the y-axis by the strip's factor. Thick ticks mark medians; the x-axis is shared across all strips within a panel. 

\begin{figure*}[pos=ht]
    \centering
    \includegraphics[width=0.9\textwidth]{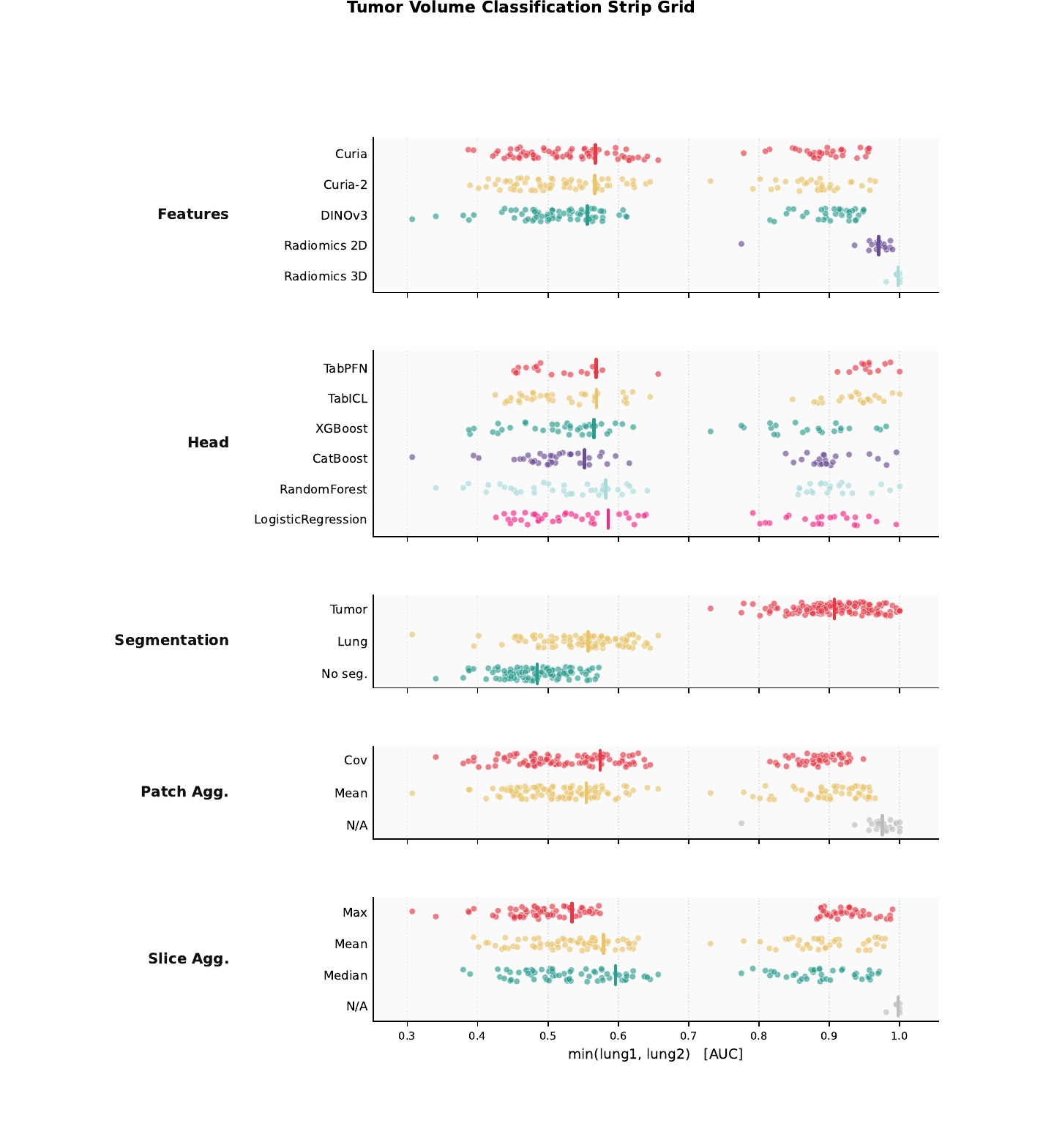}
    \caption{\textbf{Tumor volume: robustness by factor.} Metric is AUC; each dot is one configuration; higher is better; thick tick is the median. Radiomics features sit clearly to the right of all other extractors, a direct consequence of the volume label being derived from the same tumor segmentation used by radiomics; foundation-model configurations show larger spread. Segmentation choice has a pronounced effect: the tumor mask yields the strongest benefit, lung-mask outperforms no-segmentation, and no-segmentation configurations cluster at low AUC. Patch and slice aggregation choices are nearly indistinguishable.}
    \label{fig:strip_volume}
\end{figure*}

\begin{figure*}[!ht]
    \centering
    \includegraphics[width=0.9\textwidth]{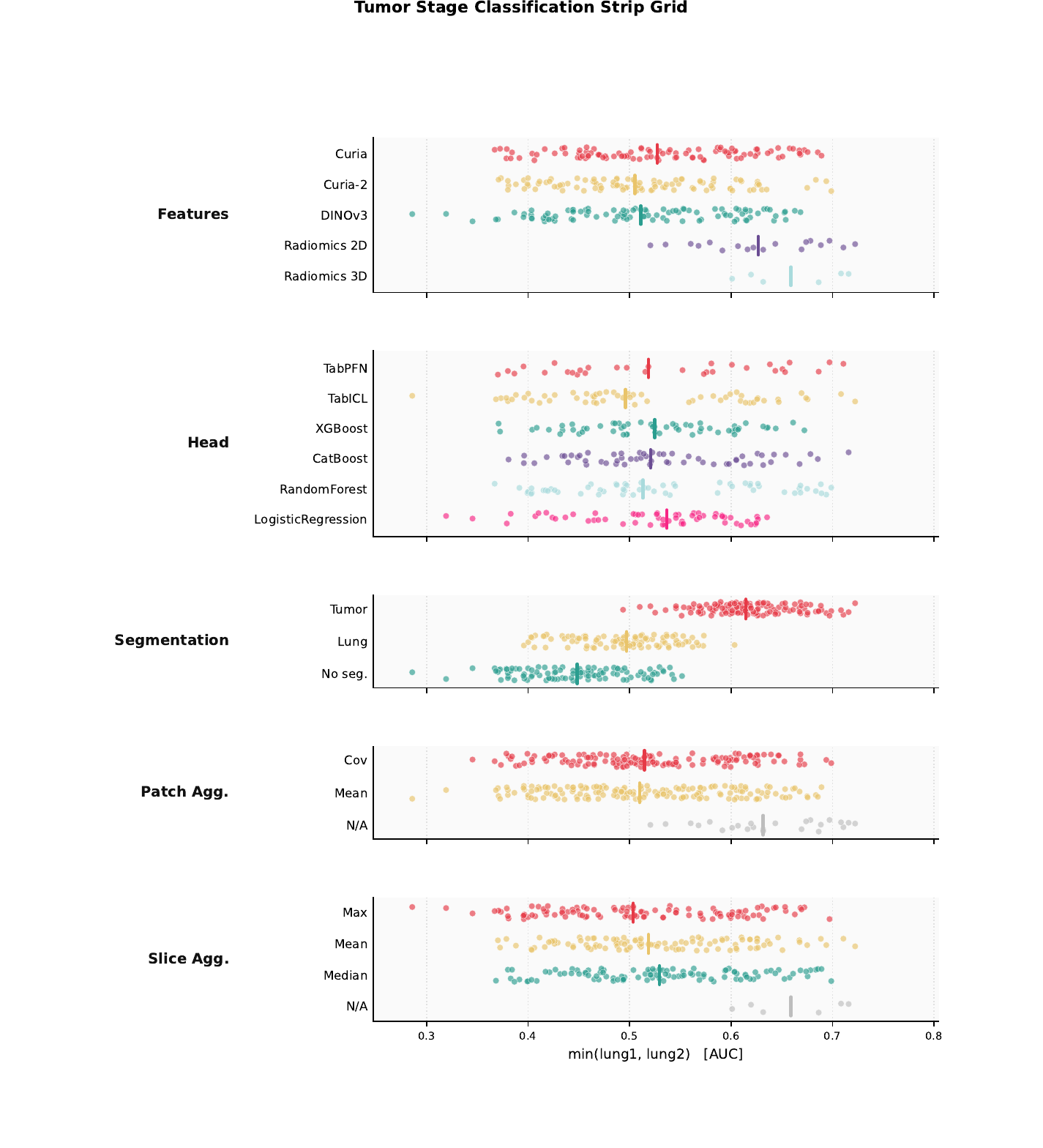}
    \caption{\textbf{Tumor stage classification: robustness by factor.} Metric is AUC; each dot is one configuration; higher is better; thick tick is the median. Radiomics 2D and 3D achieve the highest median robust scores among feature extractors, with foundation models showing comparable spread but lower medians. Segmentation is the dominant factor: the tumor mask yields a strong advantage (median clearly right of lung-mask and no-segmentation), consistent with tumor size being a defining staging criterion. Classification head and aggregation choices have moderate to negligible impact.}
    \label{fig:strip_tstage}
\end{figure*}

\begin{figure*}[!ht]
    \centering
    \includegraphics[width=0.9\textwidth]{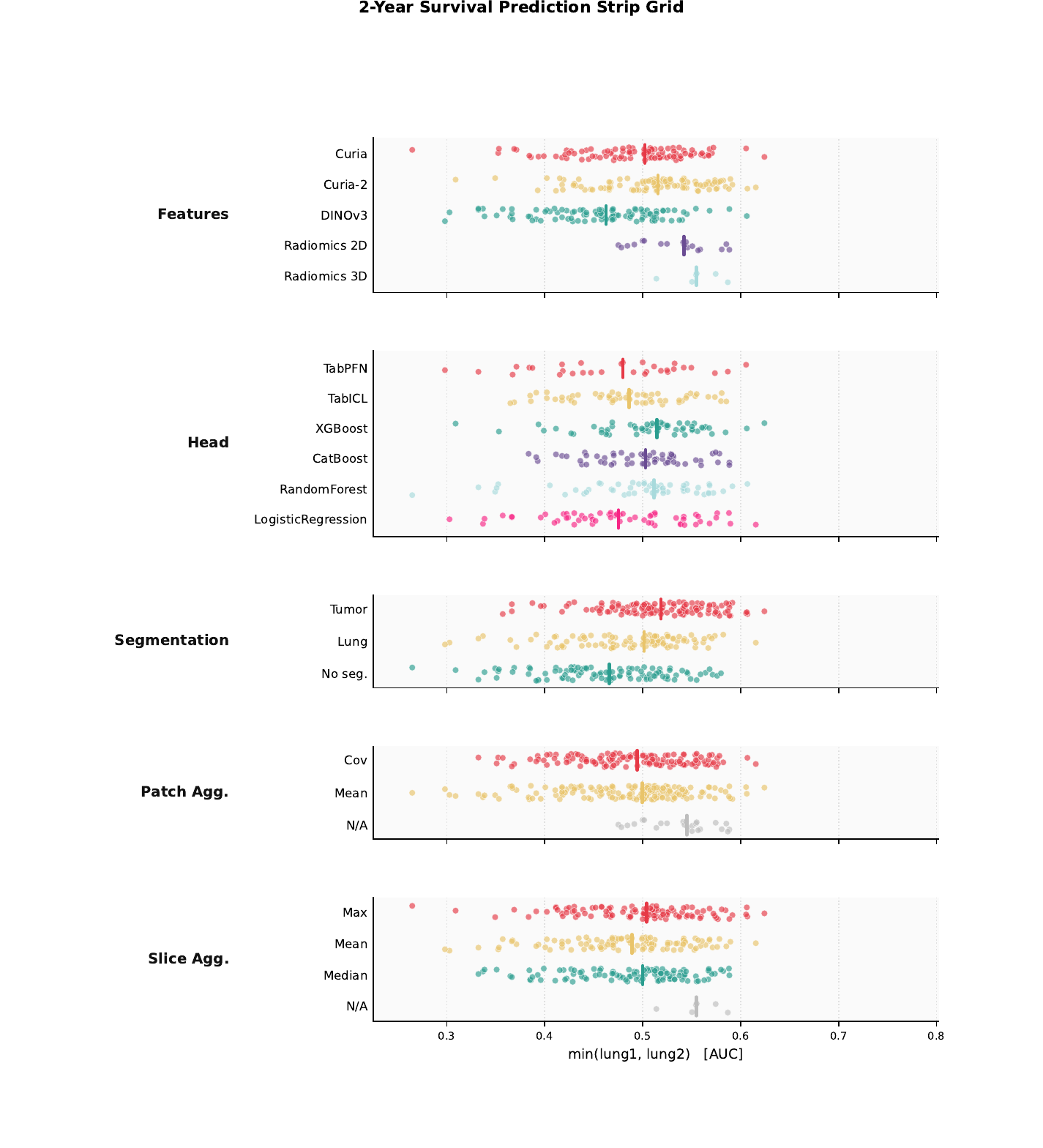}
    \caption{\textbf{2-year survival prediction: robustness by factor.} Metric is AUC; each dot is one configuration; higher is better; thick tick is the median. Radiomics achieves the highest median robust score among feature extractors, though foundation models span a wider range and reach higher peak scores. The tumor mask yields the highest median among segmentation choices; lung-mask configurations sit between tumor mask and no-segmentation. Classifier choice introduces some spread. Patch and slice aggregation choices are nearly indistinguishable.}
    \label{fig:strip_survival}
\end{figure*}

\begin{figure*}[!ht]
    \centering
    \includegraphics[width=0.9\textwidth]{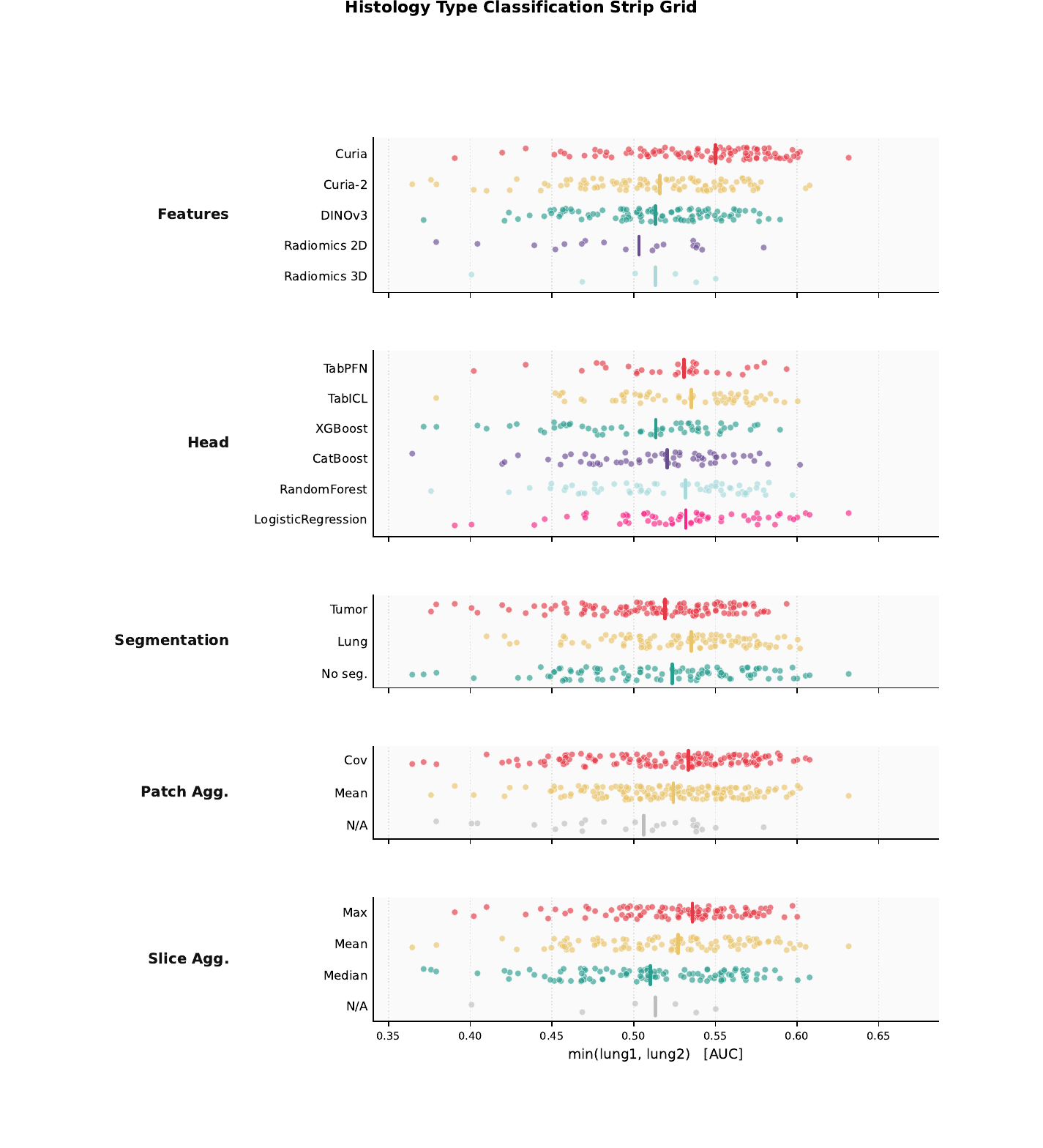}
    \caption{\textbf{Histology type classification: robustness by factor.} Metric is AUC; each dot is one configuration; higher is better; thick tick is the median. Curia achieves the highest median robust AUC among feature extractors, with Curia-2 and DINOv3 close behind; radiomics shows lower medians for this task. The three segmentation choices yield similar median scores, with lung segmentation achieving a slightly higher median, indicating that histology type classification is less sensitive to spatial masking than staging tasks. Classifier and aggregation choices have limited impact.}
    \label{fig:strip_histology}
\end{figure*}

\begin{figure*}[!ht]
    \centering
    \includegraphics[width=0.9\textwidth]{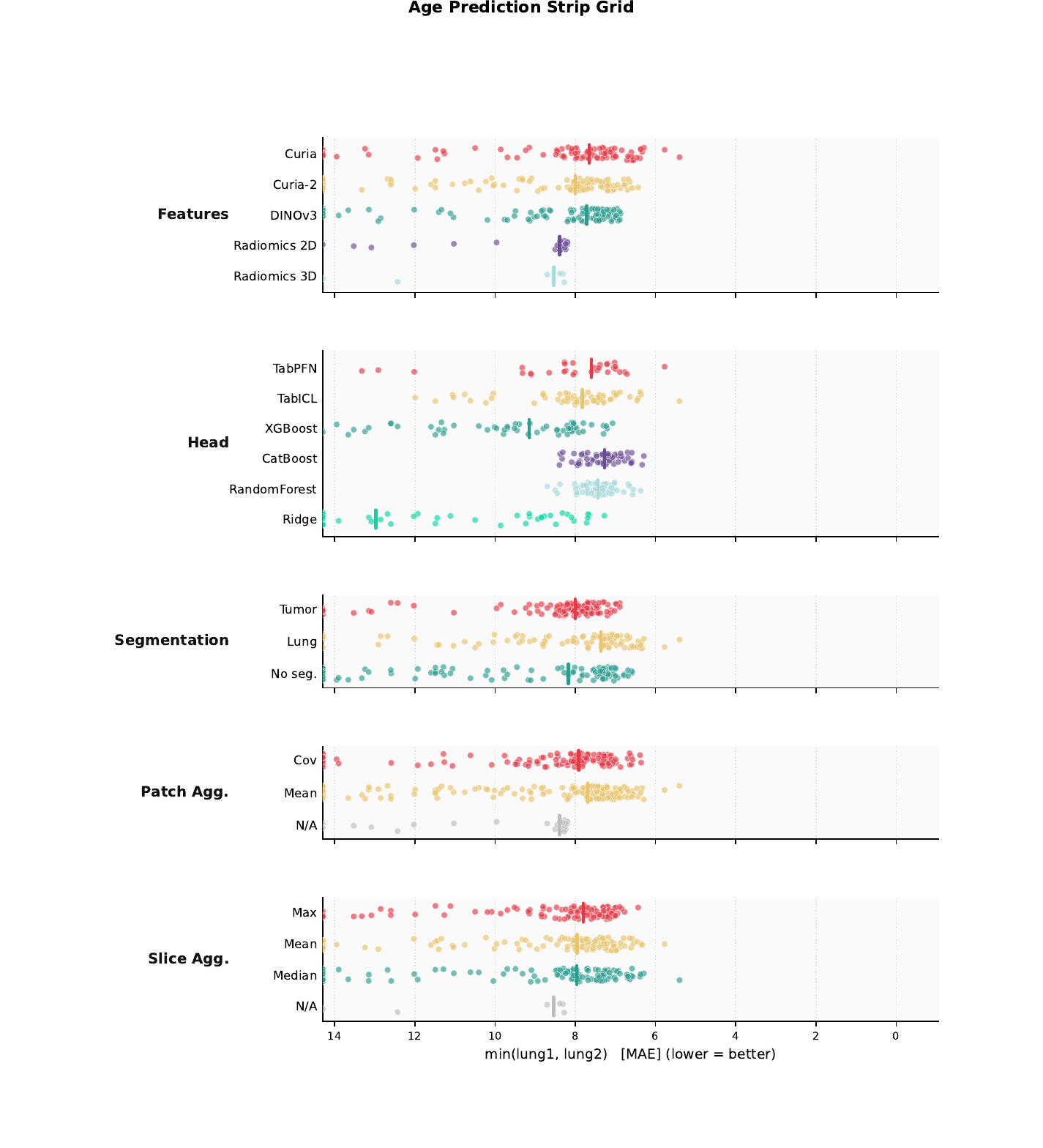}
    \caption{\textbf{Age prediction: robustness by factor.} Metric is MAE (years); each dot is one configuration; lower is better, so the x-axis is reversed; thick tick is the median. Ridge regression performs poorly relative to all other heads by a large margin, its median robust MAE substantially exceeds that of tree-based and boosting methods, driving the large classifier $\Delta$ of 9.451\,years reported in Table~\ref{tab:fi_summary}. TabICL achieves the best robust MAE, followed by TabPFN and tree-based methods. Radiomics 2D and 3D achieve higher median MAE than foundation-model extractors. Segmentation and aggregation choices have limited impact on robust performance.}
    \label{fig:strip_age}
\end{figure*}

\clearpage

\bibliographystyle{cas-model2-names}


\end{document}